\documentclass{article}
\usepackage{spconf,amsmath,graphicx}
\usepackage{algorithm}
\usepackage{algorithmic}
\usepackage{multirow}
\usepackage{bm}
\usepackage{xcolor}
\usepackage{upgreek}

\title{Motif-topology and Reward-learning improved Spiking Neural Network for Efficient Multi-sensory Integration}
\name{Shuncheng Jia $^{1,2,\#}$,
Ruichen Zuo $^{1,4,\#}$,
Tielin Zhang $^{1,2,\#,*}$,
Hongxing Liu $^{1}$,
\thanks{$^\#$ These authors contributed equally.}
Bo Xu $^{1,2,3,*}$
\thanks{$^*$ The corresponding authors are Tielin Zhang (tielin.zhang@ia.ac.cn) and Bo Xu (xubo@ia.ac.cn). This work was supported by the National Key R\&D Program of China (No. 2020AAA0108600), the Strategic Priority Research Program of the Chinese Academy of Sciences (No. XDB32070100 and No. XDA27010404), and the Shanghai Municipal Science and Technology Major Project.}
}
\address{
$^{1}$Institute of Automation, Chinese Academy of Sciences, China\\
$^{2}$School of Artificial Intelligence, University of Chinese Academy of Sciences, China\\
$^{3}$Center for Excellence in Brain Science and Intelligence Technology, CAS, China. \\
$^4$School of Information and Electronics, Beijing Institute of Technology, China
}
\begin{document}
%
\maketitle
\begin{abstract}
Network architectures and learning principles are key in forming complex functions in artificial neural networks (ANNs) and spiking neural networks (SNNs). SNNs are considered the new-generation artificial networks by incorporating more biological features than ANNs, including dynamic spiking neurons, functionally specified architectures, and efficient learning paradigms. In this paper, we propose a Motif-topology and Reward-learning improved SNN (MR-SNN) for efficient multi-sensory integration. MR-SNN contains 13 types of 3-node Motif topologies which are first extracted from independent single-sensory learning paradigms and then integrated for multi-sensory classification. The experimental results showed higher accuracy and stronger robustness of the proposed MR-SNN than other conventional SNNs without using Motifs. Furthermore, the proposed reward learning paradigm was biologically plausible and can better explain the cognitive McGurk effect caused by incongruent visual and auditory sensory signals.
\end{abstract}
\section{Introduction}
\label{sec:intro}

Spiking neural networks (SNNs) are considered as the third generation of artificial neural network (ANNs)~\cite{maass1997networks}, which are biologically plausible at both network architectures and learning paradigms. The neurons, synapses, networks, and learning principles in SNNs are far more complex and powerful than those used in ANNs~\cite{hassabis2017neuroscience}. 

This paper highlights two important features of SNNs, which are also the most differences between SNNs and ANNs, including specific network architectures and efficient learning principles. For the architectures, specific cognitive topologies learned from evolution are highly sparse and efficient in SNNs \cite{luo2021architectures}, instead of pure densely-recurrent ones in counterpart ANNs. For the learning principles, SNNs are more tuned by biologically-plausible plasticity principles, e.g., the spike timing-dependent plasticity (STDP)~\cite{zhang2017aplasticity}, short-term plasticity (STP)~\cite{zhang2018brain} (which further includes facilitation and depression), lateral inhibition, Long-Term Potentiation (LTP), Long-Term Depression (LTD), Hebbian learning, synaptic scaling, synaptic redistribution and reward-based plasticity~\cite{abraham1996metaplasticity}, instead of by the pure multi-step backpropagation (BP) of errors in ANNs. The SNNs encode spatial information by fire rate and temporal information by spike timing, giving us hints and inspiration that SNNs are also powerful in integrating visual and auditory sensory signals. 

In this paper, we focus more on the key feature of SNNs at information representation, integration, and classification. Hence, a Motif-network and Reward-learning improved SNN (MR-SNN) is proposed and then will be verified efficient on multi-sensory integration. The MR-SNN contains at least three key advantages. First, specific Motif circuits can improve accuracy and robustness at single-sensory and multi-sensory classification tasks. Second, MR-SNN can reach a relatively little higher and lower computation cost than other state-of-the-art SNNs without Motifs. Third, the reward learning paradigm can better describe the McGurk effect \cite{Tiippana2014WhatIT}, which describes an interesting psychological phenomenon that a new but reasonable audio concept might generate as a consequence of giving incongruent visual and auditory inputs. It exhibits a biologically-like behavior by using biologically plausible learning principles.

\begin{figure*}[htbp]
\centering
\includegraphics[width=16cm]{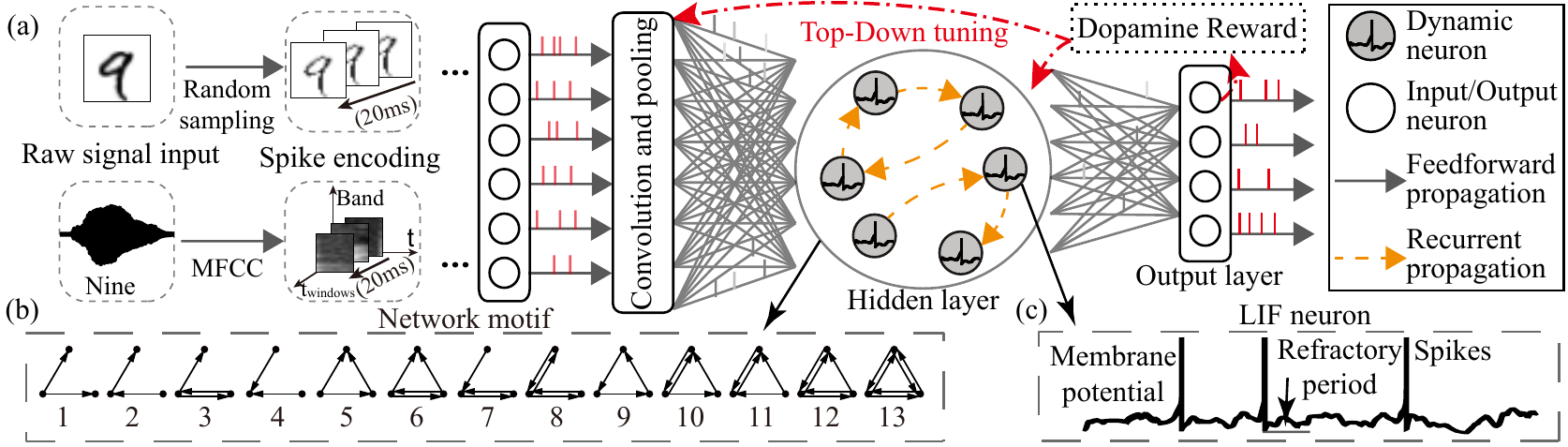}
\caption{The architecture of MR-SNN. (\textbf{a}) The architecture of MR-SNN on the multi-sensory integration task. (\textbf{b}) The example of artificial 3-node Motifs. (\textbf{c}) The spiking neuron with dynamic membrane potential.}
\label{fig_MD_SNN}
\end{figure*}

\section{Related works}
\label{sec:related}

For biological connections, the visual-haptic integration pathway~\cite{ernst2002humans}, visual-vestibular integration pathway~\cite{cheng2016distributed}, and visual-auditory integration pathway \cite{stein1989behavioral} have been identified and played important roles in opening the black box of cognitive functions in the biological brain \cite{2012Information}.

For learning paradigms, besides biologically plausible principles (e.g., STDP, STP), some efficient algorithms have been proposed, such as ANN-to-SNN conversion (i.e., directly train ANNs with BP first and then equivalently convert to SNNs)~\cite{diehl2015fast}, proxy gradient learning (i.e., replacing the non-differential membrane potential at firing threshold by an infinite gradient value) \cite{lee2016training}, and temporal BP learning (e.g., SpikeProp) \cite{bohte2002error}.

For application tasks, SNNs have shown powerful on motion planning~\cite{rueckert2016recurrent}, visual pattern recognition~\cite{zhang2018plasticity,zhang2018brain} and probabilistic inference~\cite{soltani2010synaptic}.

\section{Methods}
\label{sec:methods}

\subsection{The Motif topology}

All the 13 types of 3-node Motifs in Fig. \ref{fig_MD_SNN}(b) have been used to analyze functions in various types of biological networks \cite{2012Information}. Here, for simplicity, we transform the synaptic weights in SNNs into the range of 0 to 1 by using the Sigmoid function first and then count the Motif distribution to generate the Motif mask, named as $M_t^{r,l}$ at the recurrent layer $l$.

\subsection{The spiking neurons and recurrent architectures}

The leaky integrated-and-fire (LIF) neurons are the most simple and basic neurons for simulating biologically plausible spiking dynamics, including non-differential membrane potential and refractory period, as shown in Fig. \ref{fig_MD_SNN}(c).

The recurrent connections between LIF neurons are usually used to simulate the network-scale dynamics in SNNs. As shown in Fig. \ref{fig_MD_SNN}(a), we designed a four-layer SNN architecture containing visual and auditory input encoding, multi-sensory integration in a recurrent hidden layer, and readout layer. The synaptic weights between neurons in hidden layers are adaptive, while predefined Motif masks decide the connections between neurons. The membrane potentials in hidden layers are the integration of feedforward potential and recurrent potential, shown as follows:

\begin{equation}
    \left\{\begin{array}{l}
    \begin{matrix}
        S(t) = S^f(t) + S^r(t)
    \end{matrix}\\
    \begin{matrix}
        V_i(t) = V_i^f(t) + V_i^r(t)
    \end{matrix}\\
    \begin{matrix}
        C\frac{dV_i^f(t)}{dt}=g(V_i(t)-V_{rest})(1-S(t))+\sum_{j=1}^NW^f_{i,j}X_j
    \end{matrix}\\
    \begin{matrix}
        C\frac{dV_i^r(t)}{dt}=\sum_{j=1}^NW^r_{i,j}S(t)\cdot M_{t}^{r,l}
    \end{matrix}\\
    \end{array}\right.
    \textbf{,}
    \label{equa_recurrent_SNN}
\end{equation}

where $C$ is the capacitance, $S(t)$ is the firing flag at timing $t$, $V_i(t)$ is the membrane potential of neuron $i$ that incorporates feed-forward $V_i^f(t)$ and recurrent $V_i^r(t)$, $V_{rest}$ is the resting potential, $W_{i,j}^f$ is the feed-forward synaptic weight from the neuron $i$ to the neuron $j$, $W_{i,j}^r$ is the recurrent synaptic weight from the neuron $i$ to the neuron $j$. $M_{t}^{r,l}$ is the mask that incorporates Motif topology to influence the feedforward propagation further. The historical information is stored in the forms of recurrent membrane potential $V_i^r(t)$, where spikes are generated after potential reaching a firing threshold, shown as follows:

\begin{equation}
    \left\{\begin{array}{l}
    V_{i}^{f}(t)=V_{reset}, S^{f}(t)=1 \quad if (V_{i}^{f}(t)=V_{th}) \\
    V_{i}^{r}(t)=V_{reset}, S^{r}(t)=1 \quad if \left(V_{i}^{r}(t)=V_{t h}\right) \\
    S^{f}(t)=1 \quad i f\left(t-t_{s^{f}}<\tau_{r e f}, t \in\left(1, T_{1}\right)\right) \\
    S^{r}(t)=1 \quad i f\left(t-t_{s^{r}}<\tau_{r e f}, t \in\left(1, T_{2}\right)\right)
    \end{array}\right.
    \textbf{,}
    \label{equa_SnnUnit1}
\end{equation}

where $V_i^f(t)$ is the feed-forward membrane potential, $V_i^r(t)$ is the recurrent membrane potential, $S^f(t)$ and $S^r(t)$ are spike flags of feed-forward and recurrent membrane potentials, respectively, $V_{reset}$ is reset membrane potential.

\subsection{The local principle of gradient approximation}

The membrane potential at the firing time is a non-differential spike, so local gradient approximation (pseudo-BP)  \cite{Tuningzhang2020} is usually used to make the membrane potential differentiable by replacing the non-differential part with a predefined number, shown as follows:

\begin{equation}
    Grad_{local}=\frac{\partial S_i(t)}{\partial V_i(t)}=\left\{\begin{array}{cc}
1 & i f\left(\left|V_i(t)-V_{t h}\right|<V_{win}\right) \\
0 & else
\end{array}\right.
\textbf{,}
\label{equa_local}
\end{equation}

where $Grad_{local}$ is the local gradient of membrane potential at the hidden layer, $S_i(t)$ is the spike flag at neuron $i$, $V_i(t)$ is the membrane potential of neuron $i$, $V_{th}$ is the firing threshold. This approximation makes the membrane potential $V_i(t)$ differentiable at the spiking time between an upper bound of $V_{th}+V_{win}$ and a lower bound of $V_{th}-V_{win}$.

\subsection{The global principle of reward learning}

The reward propagation has been proposed in our previous work \cite{Tuningzhang2020}, where the reward signal is directly given to all hidden neurons without layer-to-layer backpropagation, shown as follows: 

\begin{equation}
\left\{\begin{array}{l}
Grad_{R}=B_{rand}^{f,l}\cdot R_{t}-h^{f,l} \\
\Delta W_{t}^{f,l}=-\eta^{f}(Grad_{R}) \\
\Delta W_{t}^{r,l}=-\eta^{r}\left(Grad_{t+1}+Grad_{R}\right)\cdot M_{t}^{r,l}
\end{array}\right.\text{,}
\label{equa_r}
\end{equation}

where $h^{f,l}$ is the current state of layer $l$, $R_t$ is the predefined reward for current input signal. A predefined random matrix $B_{rand}^{f,l}$ is designed to generate the reward gradient $Grad_{R}$. $W_t^{f,l}$ is the synaptic weight at layer $l$ in feed-forward phase, $\Delta W_t^{r,l}$ is the recurrent-type synaptic modification at layer $l$ which is defined by both $Grad_{R}$ by reward learning and $Grad_{t+1}$ by iterative membrane-potential learning \cite{werbos1990backpropagation}. The $M_{t}^{r,l}$ is the mask that incorporates Motif topology to further influence the propagated gradients.

\section{Experiments}
\label{sec:experiment}

\subsection{Visual and auditory Datasets}

The MNIST dataset~\cite{lecun1998mnist} was selected as the visual sensory dataset, containing 70,000 28$\times$28 one-channel gray images of handwritten digits from zero to nine.
Among them, 60,000 images are selected for training, while the remaining 10,000 ones are left for testing. The TIDigits dataset \cite{RN798} was selected as the auditory sensory dataset, containing 4,144 spoken digit recordings from zero to nine, corresponding to those in the MNIST dataset. Each recording was sampled as 20KHz for around 1 second. Some examples are shown in Fig. \ref{fig_MD_SNN}(a).

\subsection{Experimental configurations}

We built the SNN in Pytorch and trained on TITAN Xp GPU. The network architectures for MNIST and TIDigits are the same, containing one convolutional layer (with a kernel size of 5$\times$5), one full-connection or integrated layer (with 200 LIF neurons), and one output layer (with ten output neurons). The capacitance $C$ is 1$\mu$ F/cm$^2$, conductivity $g$ is 0.2 nS, time constant $\tau_{ref}$ is 1 ms, resting potential $V_{rest}$ is equal to reset potential $V_{reset}$ with 0 mV. The learning rate is $1e$-$4$, the firing threshold $V_{th}$ is 0.5 mV, the simulation time $T$ is set as 28 ms, the gradient approximation range $V_{win}$ is 0.5 mV.

As shown in Fig. \ref{fig_MD_SNN}(a), before being given to the input layer, the raw input signals were encoded to spike trains first by random sampling (for that in spatial image data) or by the temporal encoding of MFCC \cite{logan2000mel} (for that in temporal auditory data). The encoding aimed to convert the input to spike trains by comparing each number with a random number generated from Bernoulli sampling at each time slot of time window $T$.

\begin{algorithm}
\footnotesize
\caption{The MR-SNN algorithm.}
\label{algorithm1}
\begin{algorithmic}
\STATE 1. Initialize the network by resetting weights and all related parameters.
\STATE 2. Encode to spike trains from the raw numbers in datasets.
\STATE 3. Train the proposed MR-SNN on single-sensory datasets.
\STATE 3.1 Learn synaptic weights $w_{ij}$ and Motif masks $M_t^{r,l}(s)$, $M_t^{r,l}(t)$ of networks with pseudo-BP and reward-learning in two datasets, respectively.
\STATE 3.2 Save the Motif masks during single-sensory classification, in which the spatial $M^{r,l}_{t}(s)$ and temporal $M^{r,l}_{t}(t)$ were generated from visual and auditory tasks, respectively.
\STATE 4. Synthesize the integrated masks $M^{r,l}_{t}$ from spatial and temporal masks, where $M^{r,l}_{t}=1/2\times \left(M^{r,l}_{t}(s)+M^{r,l}_{t}(t)\right)$. And retrain the $w_{i,j}$ of the network with frozen Motif mask $M^{r,l}_{t}$. 
\STATE 5.Test the performance of SNNs using these new masks in the multi-sensory classification tasks, and make comparison.
\end{algorithmic}
\end{algorithm}

\subsection{Learning Motif-topology from single-sensory tasks}

\begin{figure}[htbp]
\centering
\includegraphics[width=8cm]{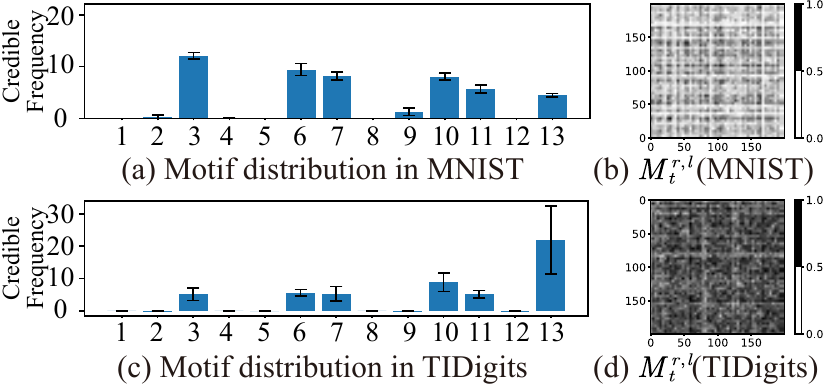}
\caption{Key Motif distributions and examples of visualization learned from the MNIST (\textbf{a}, \textbf{b}) and TIDigits (\textbf{c}, \textbf{d}) datasets. All figures are averaged over 5 repeating experiments with different random seeds. }
\label{fig_motif}
\end{figure}

The Motif distributions in visual and auditory categorizations were shown in Fig. \ref{fig_motif}(a,c), learned from different single-sensory classification tasks. We further set ``credible frequency'' by multiplying the occurrence frequency and $1-P$, where the $P$ is the P-value of a selected Motif after comparing it to 2,000 matrices that each element is uniformly and randomly distributed, indicating more credibility given by a lower P-value.

The 3rd Motif in MNIST and the 13th Motif in TIDigits are the most credibly frequent topologies. The corresponding visualization of these two types of Motif distributions is shown in Fig. \ref{fig_motif}(b,d), where the distribution on visual task is sparser than that on auditory task, which is also corresponding with the biological findings~\cite{vinje2000sparse,hromadka2008sparse}. The following sections will verify these key Motif connections playing important roles in improving accuracy and robustness during spatial and temporal classifications.

\subsection{Motif-topology for stronger robust network}

As shown in Table \ref{tab_acc}, the higher accuracies than SVM demonstrated the capacity of SNNs for dealing with sensory information. Furthermore, the experimental results showed that the SNNs using feedforward and Motif-topology (FF-Motif) were more robust than those using the feedforward connection (FF). The standard SNNs-Motif with dopamine learning~\cite{Tuningzhang2020} reached a higher accuracy than other algorithms, with an improvement of around 0.03\%, 0.43\%, and 0.03\% for MNIST, TIDigits, and integrated sensory, respectively. After giving uniformly distributed random noise to raw input, the accuracy improvement of SNN-Motif was higher than not giving noise, reaching around 7.81\%, 7.36\%, and 0.76\% for MNIST, TIDigits, and integrated sensory, respectively. It showed that these special Motif-topologies (the third one in MNIST and the 13th one in TIDigits) also contributed to the robust computation of SNNs during classification.

\subsection{Motif-topology for better multi-sensory integration}

For the experiments of multi-sensory integration, the visual and auditory signals were simultaneously given to an SNN using integrated Motif distributions learned from single-sensory tasks. As shown in Table \ref{tab_acc}, the learning accuracies for the SNNs using single visual (98.50\%) or auditory (98.20\%) sensories were lower than that using multi sensories (99.55\%). In addition, the means of accuracies for single-sensory tasks were improved up to 0.43\% for SNNs using Motif distributions than those without using them.

\begin{table}[htb]
\footnotesize
\centering
\caption{The comparison of accuracy (\%) after giving Motif topology with ($*$) or without ($-$) 80\% additional noise to the input layers.}
\begin{tabular}{lccc}
\hline
\multicolumn{1}{c}{\textbf{Topology}} & \textbf{MNIST} & \textbf{TIDigits} & \textbf{Integration} \\ \hline
SVM\cite{Platt1998SequentialMO} & 97.92 & 94.53 & -- \\ \hline
FF($-$)\cite{Tuningzhang2020} & 98.50$\pm$0.02 & 98.20$\pm$0.14 & 99.55$\pm$0.06 \\ \hline
FF-Motif($-$) & 98.53$\pm$0.09 & 98.63$\pm$0.10 & 99.58$\pm$0.09 \\ \hline
FF($*$) & 90.55$\pm$1.17 & 83.71$\pm$2.23 & 98.37$\pm$0.09 \\ \hline
FF-Motif($*$) & 98.36$\pm$0.11 & 91.07$\pm$1.42 & 99.13$\pm$0.11 \\ \hline
\end{tabular}
\label{tab_acc}
\end{table}

We found a notable accuracy increase for the paradigms with input noises, reaching 98.36\%, 91.07\%, and 99.13\% for visual, auditory, and integrated sensory, respectively. In conclusion, the Motif-topology would improve the accuracy of all three tasks under circumstances of whether giving additional noise or not.

\subsection{Motif-topology for the explainable McGurk effect}

\begin{figure}[htbp]
\centering
\includegraphics[width=8cm]{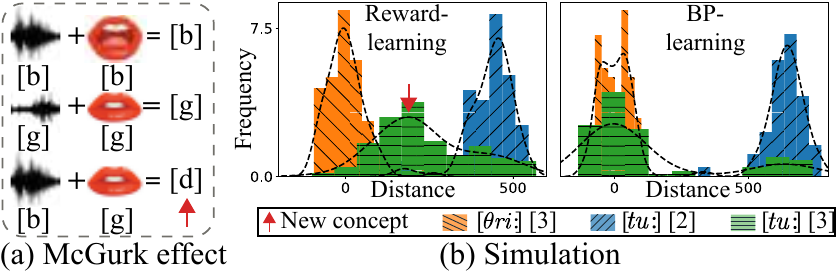}
\caption{McGurk effect and probability density distribution of feature distances at integrated layers. [$tu$:] and [$\theta ri$:] are the pronunciations of spoken two and three, [2] and [3] are the hand-digit images of two and three.}
\label{fig_integration}
\end{figure}

The McGurk effect~\cite{Tiippana2014WhatIT} describes an interesting psychological phenomenon where incongruent voice saying and face articulating will result in a new concept. For example, we believe we hear as [d] but actually voice [b] and face [g] are given, as shown in Fig.  \ref{fig_integration}(a).

We simulated this phenomenon with our proposed MR-SNN model. For simplicity, we used handwritten digits (images of [2] and [3]) and spoken digits (pronunciation of [$tu$:] and [$\theta ri$:]) to represent face articulating and voice saying, respectively. 

The two histogram figures in Fig. \ref{fig_integration}(b) were calculated to approximate the probability density of feature distance in the integrated layer (see experimental configurations for more details). We could only see two distinguished normally distributed distributions representing congruent sensory signals to SNNs using pseudo-BP, including [$\theta ri$:, 3] (orange bar) and [$tu$:, 2] (blue bar). However, besides two old concepts, a new concept (green) would generate for SNNs using reward learning and integrated Motif circuits, especially after receiving incongruent sensory signals [$tu$:, 3] (green bar). This is a small step in understanding the McGurk effect from simple algorithmic comparison. We will further discuss it in our further works.

\section{Conclusion}
\label{sec:conclusion}

We proposed a new Motif-topology and Reward-learning improved SNN (MR-SNN), exhibiting two important features. First, the Motif topology learned from spatial or temporal data could improve accuracy and robustness than standard SNNs without using Motifs. Second, with the biologically plausible reward learning, the proposed MR-SNN could simulate the McGurk effect found in cognitive integration of human brains, where the traditional SNNs using pseudo-BP would be failed. The source code of the models and experiments can be found at https://github.com/thomasaimondy/Motif-SNN.

\bibliographystyle{IEEEbib}
\bibliography{icassp}

\end{document}